\documentclass{article}

\usepackage[final]{neurips_2025}
\makeatletter
\renewcommand{\@notice}{}
\renewcommand{\ftype@noticebox}{0}
\renewcommand{\@noticestring}{}
\makeatother

\usepackage[utf8]{inputenc} 
\usepackage[T1]{fontenc}    
\usepackage{hyperref}       
\usepackage{url}            
\usepackage{multirow}
\usepackage{multicol}
\usepackage{booktabs}       
\usepackage{amsfonts}       
\usepackage{nicefrac}       
\usepackage{microtype}      
\usepackage{xcolor}         
\usepackage{wrapfig}
\usepackage{graphicx}
\usepackage{booktabs}
\usepackage{caption}
\usepackage{amssymb}
\usepackage{footnote}
\usepackage{url}
\usepackage{enumitem}
\usepackage{algorithm}
\usepackage{algorithmic}
\usepackage{amssymb}
\usepackage{listings}
\usepackage{tabularx}
\usepackage {pifont} 
\usepackage{bm}
\usepackage{tablefootnote}
\usepackage{mathtools}
\usepackage{multicol}
\usepackage{bbding}
\newcommand{\ptitle}[1]{\noindent\textbf{#1}\hspace{5pt}}
\usepackage{pifont}
\title{Unlocking Hidden Potential in Point Cloud Networks with  Attention-Guided Grouping-Feature Coordination}

\author{%
 Shangzhuo Xie, Qianqian Yang \\
College of Information Science and Electronic Engineering\\
  Zhejiang University\\
  Hangzhou, China \\
  \texttt{\{sz\_xie,qianqianyang20\}@zju.edu.cn} \\
}

\begin{document}

\maketitle

\begin{abstract}
Point cloud analysis has evolved with diverse network architectures, while existing works predominantly focus on introducing novel structural designs. 	However, conventional point-based architectures - processing raw points through sequential sampling, grouping, and feature extraction layers - demonstrate underutilized potential. We notice that substantial performance gains can be unlocked through strategic module integration rather than structural modifications. In this paper, we propose the \emph{Grouping-Feature Coordination Module} (GF-Core), a lightweight separable component that simultaneously regulates both grouping layer and feature extraction layer to enable more nuanced feature aggregation. Besides, we introduce a self-supervised pretraining strategy specifically tailored for point-based inputs to enhance model robustness in complex point cloud analysis scenarios. On ModelNet40 dataset, our method elevates baseline networks to \textbf{94.0}\% accuracy, matching advanced frameworks' performance while preserving architectural simplicity. On three variants of the ScanObjectNN dataset, we obtain improvements of \textbf{2.96}\%, \textbf{6.34}\%, and \textbf{6.32}\% respectively.
\end{abstract}

\section{Introduction}

\begin{wrapfigure}{tr}{7.1cm}
    \centering
    \vspace{-2mm}
    \includegraphics[width=0.507\textwidth]{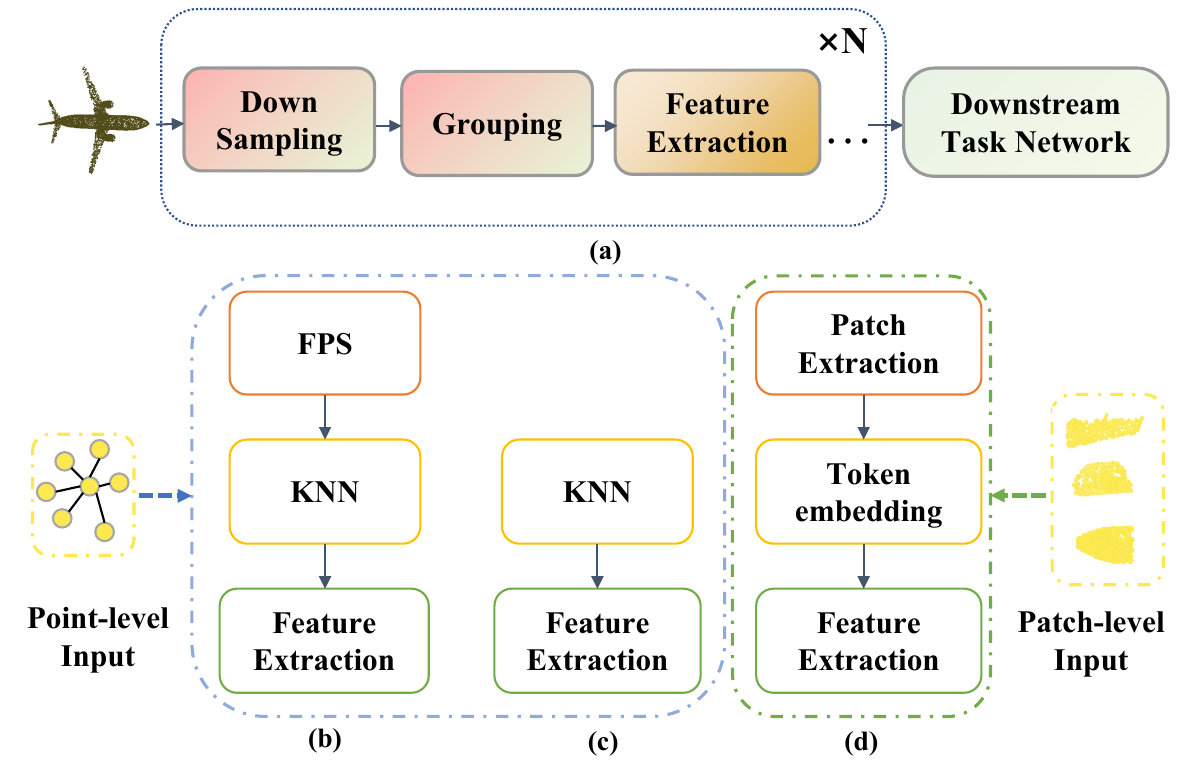}
    \vspace{-2mm}
    \caption{Point cloud processing paradigms: (a) Standard: iterative downsampling-grouping-Feature extraction pipeline, (b) Static Grouping: furthest point sampling(FPS) and K-nearest neighbors(KNN), with fixed neighborhoods based on geometry, (c) Dynamic Grouping: continuous KNN graph construction, (d) Patch Grouping: patch embedding followed by attention/state-space model}
    \vspace{-2mm}
    \label{fig:arch}
\end{wrapfigure}

Point clouds, which represent raw 3D geometry as unordered sets of points, pose significant computational challenges due to their irregular structure and permutation invariance~\cite{pointcloud1,pointcloud2,pointcloud3}. As illustrated in Fig.~\ref{fig:arch}, modern point cloud processing frameworks typically adopt a sequential pipeline comprising downsampling, neighborhood grouping, and feature extraction. A seminal example is PointNet++~\cite{PointNet++}, which introduced hierarchical processing by iteratively applying Farthest Point Sampling (FPS), constructing local neighborhoods, and aggregating point features via multilayer perceptrons (MLPs). Subsequent studies~\cite{PointCNN,RS-CNN,pointmlp,mixer,Achlioptas2017RepresentationLA} further enhanced this paradigm by refining local geometric reasoning within static grouping strategies. Meanwhile, dynamic approaches such as DGCNN~\cite{DGCNN} redefined grouping through adaptive KNN-based graph construction, a direction later extended by works~\cite{KPConv,Curvenet} to incorporate advanced geometric priors. 


Recent works have explored paradigm shifts in architectural design for point cloud analysis. Transformer-based approaches~\cite{Point-BERT,Point-MAE,Point-M2AE,MAE3D} and Mamba-driven methods~\cite{Pointmamba,mamba3D,Zhang2024PointCM}  have demonstrated strong performance by reinterpreting point clouds as sequences of patch-based tokens processed via attention mechanisms or state-space models. While effective, these approaches fundamentally depart from the data's native structure: partitioning raw points into fixed patches can obscure fine-grained geometric relationships that classical point-based frameworks are designed to preserve explicitly.


Transformer- and Mamba-based approaches, while achieving impressive performance, often introduce considerable architectural complexity—resulting in increased inference latency in point cloud analysis. This raises a critical question: Is complexity a necessary driver of progress? PointMLP~\cite{pointmlp} matches transformer-level accuracy using refined MLP operators and hierarchical feature propagation, illustrating that lightweight enhancements to classical pipelines can deliver competitive performance without architectural overhauls. Similarly, PointNeXt~\cite{PointNeXt} demonstrates that conventional architectures, when paired with adaptive training protocols and optimized local geometric aggregation, can surpass even state-of-the-art attention-based models. These advancements suggest that conventional point-based architectures, which preserve geometric relationships within the raw point input, retain substantial untapped potential to advance point cloud analysis without increasing model complexity.

To unlock the potential of classical point-based frameworks such as PointNet++, we revisit their design philosophy to address two fundamental limitations: (1) information loss caused by max-pooling during feature aggregation, and (2) suboptimal neighborhood selection introduced by fixed k-NN grouping. These issues can lead to the loss of critical geometric details, particularly in fine-grained structures. To mitigate these limitations, we propose \textbf{Grouping-Feature Coordination Module (GF-Core)}—a graph-aware architecture that integrates the inherent graph topology of point clouds with adaptive attention mechanisms. Within the feature extraction stage, GF-Core computes attention weights between neighborhood centroids and their neighboring points by jointly leveraging geometric and feature cues. It then performs weighted fusion with edge features extracted from the backbone network to obtain high-quality local representations. Simultaneously, the computed weights are organized into a weight matrix, which propagates neighborhood information to subsequent grouping layers. This enables GF-Core to refine neighborhood selection by jointly considering spatial distributions and learned feature affinities.


To further enhance the module's ability to integrate local neighborhood information, we introduce a self-supervised pre-training strategy based on grouping perturbation. This strategy applies coordinated point-level perturbations and masking operations across multiple layers. Subsequently, we perform feature comparison between the perturbed and original point clouds using the Barlow Twins loss~\cite{barlowtwins}, aiming to maximize cross-view feature consistency while minimizing redundancy under adversarial disturbances. This approach significantly improves the network's capacity to learn discriminative point-level features in a robust and data-efficient manner.

The proposed GF-Core module and self-supervised learning mechanism are model-agnostic and can be integrated into any point-based architecture to enhance point cloud analysis performance.  Experimental results show that our approach boosts DGCNN to 94.0\% accuracy (+1.1\%) on the ModelNet40 classification benchmark with only 2.1M parameters, surpassing most transformer-based architectures. For real-world robustness, our method achieves significant accuracy improvements of +6.34\% and +6.32\% on the challenging OBJ\_BG and PB\_T50\_RS variants of ScanObjectNN, respectively, while using DGCNN as the backbone.

\section{Related work}

\ptitle{Point cloud analysis.} The advancement of point cloud understanding has been propelled by continuous innovations in geometric feature learning. Early breakthroughs began with PointNet~\cite{pointnet}, which introduced permutation-invariant processing via symmetric max-pooling, laying the groundwork for learning directly from raw point sets. Its successor, PointNet++~\cite{PointNet++}, extended this paradigm by incorporating hierarchical feature abstraction through iterative farthest point sampling and multi-scale grouping, enabling context-aware representation learning. Building on these foundations, subsequent approaches enriched local geometric modeling through dynamic graph convolutions~\cite{DGCNN,PAConv}, which adaptively update neighborhood relationships across network layers, and continuous convolution kernels~\cite{PointConv,KPConv,Mao2019InterpolatedCN,Zhou2021AdaptiveGC}, which overcome the limitations of discrete grids via differentiable density estimation. Further expanding this landscape, geometric primitive-aware operations such as SpiderCNN~\cite{SpiderCNN} captured parameterized surface variations, while PointSIFT~\cite{PointSIFT} encoded orientation-sensitive features using sequential axis-aligned rotations. In parallel, sparse convolutional frameworks~\cite{minkowski} were developed to efficiently handle large-scale point clouds using high-dimensional spatiotemporal hashing. More recently, architectural innovation has shifted toward transformer-based models~\cite{PCT1,PCT2,PCT3,PCT4,Wu2022PointTV}, which adapt self-attention mechanisms to capture long-range dependencies, and Mamba-based models~\cite{Pointmamba,mamba3D,Zhang2024PointCM}, which leverage state-space representations for efficient sequence modeling with linear computational complexity.

\ptitle{Attention mechanism for point cloud analysis.} Attention mechanisms enable dynamic feature weighting for point cloud analysis and hold strong potential for improving performance. Early adaptations of self-attention~\cite{PCT1,PCT2} directly transplanted transformer architectures into the 3D domain, achieving global context modeling through pairwise similarity computations across all points. However, these methods suffer from quadratic computational complexity, which limits their scalability to dense point clouds. To address this, numerous works~\cite{PointASNL,APCEF,edgconv,Wu2023AttentionBasedPC} have improved efficiency through adaptive sampling and edge-aware attention strategies. In addition, graph-based methods~\cite{PAT,PANet,AGConv} and linear attention approaches~\cite{LinNet} have inspired lightweight attention mechanisms that reduce computational overhead while maintaining representational effectiveness.

\ptitle{Self-supervised learning for point cloud analysis.}  Self-supervised learning has emerged as a powerful approach for geometric representation learning, significantly reducing the dependency on costly labeled data. In point cloud analysis, self-supervised learning has evolved along two primary trajectories: contrastive learning frameworks, which emphasize feature discrimination through instance-level comparisons~\cite{Jiang2021GuidedPC,Zhang2021SelfSupervisedPO,Sanghi2020Info3DRL,Liu2021PointDL,Self-Contrastive}, and reconstruction-based approaches, which focus on recovering geometric structures via masking strategies. The contrastive paradigm, exemplified by PointContrast~\cite{pointcontrast}, promotes feature consistency by aligning geometrically transformed point cloud pairs, while CrossPoint~\cite{crosspoint} extends this concept to cross-modal alignment between 3D points and 2D projections. On the other hand, reconstruction-based approaches, such as OcCo~\cite{occo}, introduced partial-to-complete shape recovery as a precursor to modern masked autoencoding techniques. Inspired by the success of BERT-style pretraining in NLP~\cite{devlin2018bert}, recent works~\cite{Point-BERT,Point-MAE,Point-M2AE,MAE3D,Chen2023PiMAEPC,Yan20233DFP} adapt masked autoencoding to point clouds by discretizing local patches into geometric tokens and performing coordinate regression under high masking ratios. However, these reconstruction-based methods risk overemphasizing global shape recovery at the expense of fine-grained semantic understanding. In this paper, we propose a unified strategy that synergizes the strengths of both paradigms. By integrating deformation-aware contrastive objectives with masked reconstruction, our method enables efficient and discriminative feature learning at the point level.

\section{Methodology}

\begin{figure*}[!t]
	\begin{center}
		\includegraphics[width=1.0\linewidth]{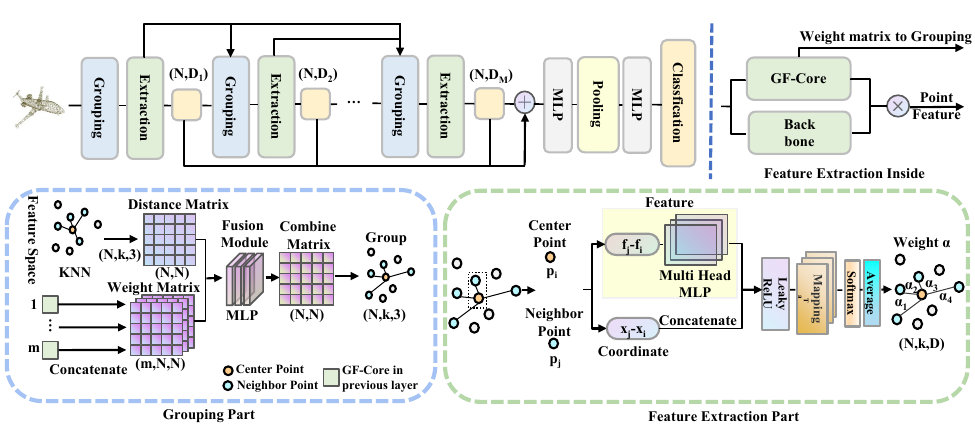}
	\end{center}
    \vspace{-4mm}
	\caption{Illustration of the point cloud analysis framework integrating the proposed \textbf{GF-Core} module with DGCNN. GF-Core is applied at two stages: feature extraction and grouping. During feature extraction, GF-Core processes centroid–neighbor pairs by jointly modeling geometric displacements and feature differences, computes attention weights via multi-head fusion, and aggregates features using softmax-normalized adaptive weighting. In the grouping stage, it dynamically refines neighborhood selection by combining learned attention matrices with spatial distance metrics. Hierarchical weight matrices are propagated through skip connections to guide multi-layer feature aggregation.}
	\label{fig:model}
    \vspace{-5mm}
\end{figure*}

We present the proposed GF-Core module and self-supervised learning mechanism in this section. Fig.~\ref{fig:model} illustrates the framework which integrates GF-Core module with DGCNN to perform poinr cloud classification. Notably, the GF-Core module is model-agnostic and can be incorporated into any point-based architecture to enhance point cloud analysis performance. The GF-Core module is applied to two steps: \textbf{Feature Extraction} and \textbf{Grouping}, which respectively regulate neighborhood feature aggregation and guide the selection of aggregated neighborhoods.

In the following, we first introduce the background of graph-based attention in Sec.\ref{GAT}, then provide a detailed description of the GF-Core module in Sec.\ref{GF-Core}, and finally present our self-supervised pretraining strategy in Sec.~\ref{pretrain}.

\subsection{Graph Attention Networks}
\label{GAT}
Graph Attention Networks (GAT)~\cite{GAT} use a mechanism where each node aggregates information from its neighbors based on an attention score. In GAT, the attention score between nodes \(i\) and \(j\) is computed using the following equation:
\begin{equation}
\alpha_{ij} = \mathrm{Softmax}\left(\mathrm{LeakyReLU} \left( \mathbf{a}^\top \cdot [\mathbf{W} \mathbf{h}_i \parallel \mathbf{W} \mathbf{h}_j] \right)\right)
\end{equation}
where \( \mathbf{h}_i \) and \( \mathbf{h}_j \) are the feature vectors of nodes \(i\) and \(j\), \( \mathbf{W} \) is a learnable weight matrix, \( \mathbf{a} \) is a learnable attention vector. The aggregated feature for node \(i\) is computed as the weighted sum of its neighbors' features:
\begin{equation}
\mathbf{h}_i' = \sum_{j \in \mathcal{N}(i)} \alpha_{ij} \mathbf{W} \mathbf{h}_j
\end{equation}
The key advantage of the graph attention mechanism lies in its ability to learn attention weights dynamically, rather than relying on fixed graph structures. First, the weight matrix \(\mathbf{W}\) projects node features into a latent space where semantic relationships can be more effectively captured. The concatenated features \([\mathbf{W}\mathbf{h}_i \parallel \mathbf{W}\mathbf{h}_j]\) are then processed by an attention vector \(\mathbf{a}\) enabling the modeling of pairwise interactions. The LeakyReLU activation introduces non-linearity into the attention computation, while the softmax function normalizes the attention scores so that they sum to one over each node’s neighborhood.

GATv2~\cite{GATv2} further improves upon GAT by adopting an additive mechanism rather than the bilinear form used in GAT. Specifically, the attention score in GATv2 is computed as:
\begin{equation}
\alpha_{ij} = \mathrm{Softmax}\left(\mathbf{a}^\top \mathrm{LeakyReLU} \left( \mathbf{W} \cdot [\mathbf{h}_i \parallel \mathbf{h}_j] \right)\right),
\end{equation}
The key difference lies in the computation order: GATv2 applies a shared linear transformation \(\mathbf{W}\) to the concatenated raw features \([\mathbf{h}_i \parallel \mathbf{h}_j]\), followed by non-linear activation, before multiplying with vector \(\mathbf{a}\). This enables the attention coefficients to dynamically adapt to specific query-node pairs, allowing the model to distinguish identical neighbor features across different queries and learn richer, more expressive attention patterns. This dynamic property is particularly valuable for point cloud analysis, where geometric relationships—such as surface normals or curvature—demand adaptive neighborhood weighting. For example, a point on a sharp edge may prioritize geometrically aligned neighbors, while one on a smooth surface may emphasize semantic cues. GATv2's dynamic attention naturally accommodates such spatially varying importance.

\subsection{Grouping-Feature Coordination Module}
\label{GF-Core}
We now present the proposed GF-Core module, which is applied at two stages of the point cloud analysis pipeline.

\ptitle{Feature Extraction Stage.} Consider \(\mathbf{p}_i\) as the center point and \(\mathbf{p}_j\) as its neighboring point, where \(\mathbf{p}_{i/j} = (\mathbf{x}_{i/j},\mathbf{f}_{i/j})\in \mathbb{R} ^{3+d}\). We have 
\begin{equation}
\mathbf{f}_{ij} = \mathbf{f}_j - \mathbf{f}_i \in \mathbb{R}^d, \quad 
\mathbf{x}_{ij} = \mathbf{x}_j - \mathbf{x}_i \in \mathbb{R}^3,
\end{equation}
where \( \mathbf{f}_{ij} \) captures feature differences in \( d \)-dimensional semantic space and \( \mathbf{x}_{ij} \) represents 3D geometric displacement. Attention mechanisms are then applied:
\begin{equation}
\mathbf{G}_h = \mathrm{MLP}_h(\mathbf{f}_{ij}) \in \mathbb{R}^{D} ,
\label{eq:parallel_process1}
\end{equation}
\begin{equation}
\mathbf{e}_{ij}^h = \mathbf{a}_h^\top \cdot \mathrm{LeakyReLU}\Big( [\mathbf{G}_h \parallel \mathbf{x}_{ij}] \Big),
\label{eq:parallel_process2}
\end{equation}
where an MLP transforms feature differences into a \( D \)-dimensional latent space in Eq.~\ref{eq:parallel_process1} , while attention energies are computed by fusing projected features with raw geometric displacements via concatenation (\( \parallel \)) and nonlinear activation in Eq.~\ref{eq:parallel_process2}. The learnable parameters \(\mathbf{a}_{h}^\top \in \mathbb{R}^{(D+3) \times D}\) adaptively balance geometric and semantic contributions. To further reduce the module’s complexity, we apply a low-rank approximation to the parameter matrix \( \mathbf{a}_h^\top\) as follows:
\begin{equation}
\mathbf{a}_h^\top = \mathbf{U}_h \mathbf{V}_h \quad
\label{eq:low_rank_decomp}
\end{equation}
where \(\mathbf{U}_h \in \mathbb{R}^{(D+3) \times R} \), \(\mathbf{V}_h \in \mathbb{R}^{R \times D}\), \(R = \lfloor D / r\rfloor\).  $r$ is a hyperparameter that controls the trade-off between performance and computational cost.

The multi-head attention weights are normalized and aggregated in two steps:
\begin{equation}
\alpha_{ij}^h = \frac{\exp(\mathbf{e}_{ij}^h)}{\sum_{k\in\mathcal{N}(i)}\exp(\mathbf{e}_{ik}^h)}, \quad 
\alpha_{ij} = \frac{1}{H}\sum_{h=1}^H \alpha_{ij}^h\in \mathbb{R}^D
\end{equation}
The first step applies neighborhood-wise softmax normalization to ensure spatial locality, while the second step averages multi-head weights to preserve perspective diversity.

Finally, neighbor features are combined through the backbone network's convolutional layer:
\begin{equation}
\mathbf{f}_i' = \sum_{j\in\mathcal{N}(i)} \alpha_{ij} \cdot \underbrace{\mathrm{Conv}(\mathbf{F}_j)}_{\mathclap{\mathrm{Base\ Network}}} \in \mathbb{R}^{D}
\label{eq:feature_propagation}
\end{equation}
Where \(\mathbf{F}_j\) and $\mathrm{Conv(\cdot)}$ vary depending on the backbone design. This approach preserves the fundamental operators of the base network while injecting adaptive attention. The attention computations enable plug-and-play integration without structural modifications.

To further refine the aggregation process, we compute a weight matrix \( \mathbf{W} \) based on the attention scores. The weight matrix is learned through an MLP applied to the attention scores:
\begin{equation}
\mathbf{W_0} = \mathbf{MLP}(\alpha)  \in \mathbb{R}^{N \times k}
\label{eq:W}
\end{equation}
Where \(\alpha\ \in \mathbb{R}^{N \times k \times D}\) represents the attention weights for all point set graphs. We then transform $\mathbf{W_0}$ into a $N \times N$ matrix  by zero-padding. The weight matrix \( \mathbf{W} \) captures interdependencies among points,enabling a more sophisticated integration of their mutual contributions and facilitating the propagation of point-dependent information across spatial scales.

\ptitle{Grouping Stage.} Traditional k-Nearest Neighbors (k-NN) grouping relies solely on Euclidean distances, limiting its ability to capture semantic relationships. GF-Core introduces a novel neighbor selection mechanism that progressively integrates learned attention patterns, allowing the model to focus on the most relevant neighbors at each layer.
The key is the concatenation of hierarchical weight matrices:
\begin{equation}
\mathbf{W}_{new} = [\mathbf{W}^1|| \mathbf{W}^2|| ...||\mathbf{W}^{m-1}],
\label{eq:concat}
\end{equation}
where $\mathbf{W}^{m-1}$ represents attention weights from the $(m-1)^{th}$ 
 layer, and $\mathbf{W}_{new}$ denotes newly learned attention patterns. As the network progresses, GF-Core concatenates the previous layers’ weight matrices with the new one. This iterative refinement ensures that at each layer, the model’s focus on relevant neighbors becomes increasingly precise, allowing the network to better capture the point cloud’s structure.

The feature distance $\mathbf{D}_{feature}^m$ between two points at the $m^{th}$ layer is computed using a pairwise feature distance measure and is subsequently refined by the fusion module to identify the top $k$ most relevant neighboring points:
\begin{equation}
\mathbf{D}_{combine}^m = \mathrm{Fusion}\left([\mathbf{D}_{feature}^m||\mathbf{W}_{new} \right])
\label{eq:fusion}
\end{equation}
\begin{equation}
\mathcal{N}_k(\mathbf{p_i}) = \mathop{\mathrm{arg\,top}_k}\limits_{j=1,...,N} \left( -\mathbf{D}_{combine}^m[i,j] \right).
\label{eq:neighbor_selection}
\end{equation}
We note that since no attention weight matrix $\mathbf{W}$ is available in the first layer, $\mathbf{D}_{combine}^{1}$ defaults to the standard k-NN distance for neighbor selection.

\begin{figure*}[!t]
	\begin{center}
		\includegraphics[width=1\linewidth]{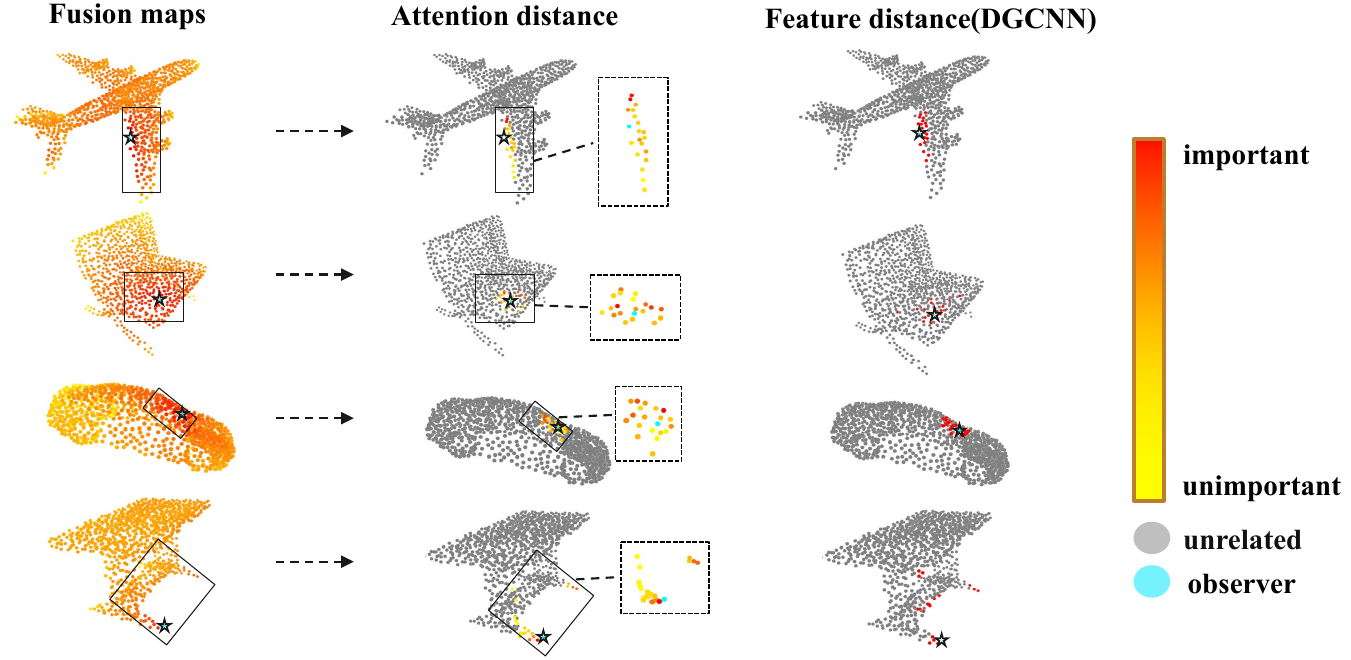}
	\end{center}
        \vspace{-4mm}
	\caption{\textbf{Visualization of point correlations by different components.} (1) Fusion maps: Inter-point correlations derived from fusion module in the grouping layer, i.e., $\mathbf{D}_{combine}^m$; (2) GF-Core attention: Learned inter-point attention weights, i.e., $\mathbf{W}^{m}$; (3) DGCNN: Learned edge connections by conventional graph convolution.
From yellow to red indicates ascending relational importance.}
	\label{fig:vis}
    \vspace{-5mm}
\end{figure*}

As shown in Fig.~\ref{fig:vis}, GF-Core effectively prioritizes critical neighboring points through attention-guided coordination, focusing on semantically relevant points that may be geometrically distant.


\subsection{Self-Supervised Pretraining via Grouping Perturbation}
\label{pretrain}
The proposed self-supervised pretraining mechanism introduces dynamic grouping perturbation within each neighborhood group $\mathcal{N}_k(\mathbf{p})$ during the grouping process:
\begin{equation}
\tilde{\mathcal{N}}_k(\mathbf{p}) = \mathrm{Shuffle}\Big( 
    \{\mathbf{x}_j, \mathbf{f}_j\}_{j=1}^{k-M} 
    \cup \{\mathbf{x}_m, \theta\}_{m=1}^M 
    \cup \{\tilde{\mathbf{x}}_a, \tilde{\mathbf{f}}_a\}_{a=1}^A \Big)
\end{equation}
where \(\tilde{\mathbf{x}}_a \sim U[\mathbf{x}_{\min}-\epsilon, \mathbf{x}_{\max}+\epsilon]\),
\(\tilde{\mathbf{f}}_a \sim \mathcal{N}(0,\sigma^2\mathbf{I})\).
This local perturbation operates in three stages: (1) Randomly mask $M$ points by replacing their features with learnable parameters $\theta$, (2) Inject $A$ noisy points within the expanded neighborhood sphere $\mathcal{B}(\mathbf{p},r{+}\epsilon)$, and (3) Shuffle the order of points. 
We then extend Barlow Twins~\cite{barlowtwins} with a geometric-semantic alignment objective:
\begin{equation}
\begin{aligned}
\mathbf{C} &= \mathrm{Norm}(\mathbf{Z}_{\text{clean}}^\top\mathbf{Z}_{\text{noisy}}) \in \mathbb{R}^{d\times d} 
\end{aligned}
\end{equation}
\begin{equation}
\begin{aligned}
\mathcal{L}_{\text{align}} &= \sum_i(1-C_{ii})^2 + \lambda\sum_{i eq j}C_{ij}^2
\end{aligned}
\label{eq:barlow}
\end{equation}
where $d$ is the feature dimension. The cross-correlation matrix $\mathbf{C}$ captures feature co-activation patterns between clean and noisy views: diagonal terms encourage invariance under perturbations, while off-diagonal terms penalize redundancy across feature dimensions.

In practical,we use multi-scale consistency learning coordinates local and global understanding:
\begin{equation}
\begin{aligned}
\mathcal{L}_{\text{total}} = \alpha \mathcal{L}_{\text{align}}(f_{clean},f_{noisy}) + (1 - \alpha) \mathcal{L}_{\text{align}}(F_{clean},F_{noisy})
\end{aligned}
\label{eq:loss_total}
\end{equation}
$f$ denotes the per-point feature obtained after all encoding layers, $F$ represents the global feature aggregated from all points. 
The coordination between these two features effectively enhances the network's capability in both global and local point cloud representation, the trade-off between local detail preservation ($\alpha \uparrow$) and global context modeling ($\alpha \downarrow$) is governed by the scaling factor $\alpha$.

\section{Experiments}

\subsection{Experimental Setting}
we meticulously configure the network architecture and hyperparameters to align precisely with the backbone network~\cite{DGCNN,PointNet++}, ensuring a consistent foundation for evaluation. Pre-training is conducted on the ShapeNet~\cite{ShapeNet} dataset, which comprises about 51,000 models spanning 55 categories. We set the input point cloud size to 1,024 points and configure the output feature dimension to 1,024. For GF-Core, we employ two attention heads, with the low-rank decomposition parameter $r$ = 8 for the \( \mathbf{a}_h^\top\). The original k-nearest neighbor parameter is retained at $k$ = 20. Across the four-layer network, we apply a progressive masking strategy with masked point counts of (6, 10, 12, 16) and perturbation point counts of (2, 4, 6, 8). The balancing coefficient $\alpha$ tailored to downstream tasks, set to 0.97 for classification and 0.98 for segmentation. During training, we use a batch size of 32 and adopt the AdamW~\cite{adamw} optimizer with an initial learning rate of 0.001 and a weight decay of 10-4, training for 200 epochs. For fine-tuning, the backbone network’s learning rate is reduced to 0.0005, while the downstream task heads use a learning rate of 0.001. All experiments are performed on NVIDIA RTX 3090 GPUs.
\begin{table}[ht]
\vspace{-2mm}
\caption{Classification results on ModelNet40 dataset. We compared representative models from various architecture. Param. denotes the number of tunable parameters during training.* represents \textbf{G} using the base network's feature extraction network.}
\centering
\begin{tabular*}{\textwidth}{@{\extracolsep{\fill}}l c l c c}
\toprule
Methods & Pre-trained & Architecture & Param. (M)  & Acc. \\ \hline
PointNet~\cite{pointnet}         & ×                    & -                     & 3.5    & 89.2 \\ 
PointNet++~\cite{PointNet++}       & ×                    & -                     & 1.5  & 90.7 \\ 
DGCNN~\cite{DGCNN}            & ×                    & -                     & 1.8    & 92.9 \\ 
PointCNN~\cite{PointCNN}         & ×                    & -                     & 0.6    & 92.5 \\ 
PointMLP~\cite{pointmlp}         & ×                    & -                     & 12.6 & 94.1 \\ 
DeLA~\cite{DELA}             & ×                    & -                     & 5.3    & 94.0 \\ \hline
PCT~\cite{PCT2}              & ×                    & Transformer           & 2.9    & 93.2 \\ 
Point-BERT~\cite{Point-BERT}       & \checkmark                    & Transformer           & 22.1   & 93.2 \\ 
Point-MAE~\cite{Point-MAE}        & \checkmark                    & Transformer           & 22.1   & 93.2 \\ 
Point-M2AE~\cite{Point-M2AE}       & \checkmark           & Transformer           & 12.8   & 93.4 \\ 
MAE3D~\cite{MAE3D}            & \checkmark           & Transformer           & 42.4   & 93.4 \\ 
Mamba3D~\cite{mamba3D}          & \checkmark           & Mamba                 & 16.9   & \textbf{94.7} \\ 
PointMamba~\cite{Pointmamba}       & \checkmark           & Mamba                 & 12.3   & 93.6 \\ \hline
Ours (PointNet++)* & \checkmark                    & -                     & 1.6    & 92.1\textcolor{red}{(+1.4)}\\ 
Ours (PointNet++) & \checkmark                    & -                     & 1.8    & 92.5 \textcolor{red}{(+1.8)} \\ 
Ours (DGCNN)*     & \checkmark                    & -                    & 2.0    & 93.7\textcolor{red}{(+0.8)} \\ 
Ours (DGCNN)     & \checkmark                    & -                    & 2.1   & 94.0 \textcolor{red}{(+1.1)} \\ 
Ours (PointMLP)*     & \checkmark                    & -                    & 12.9    & 94.5\textcolor{red}{(+0.4)} \\ 
Ours (PointMLP)     & \checkmark                    & -                    & 13.1    & 94.4\textcolor{red}{(+0.3)} \\ 
Ours-elite (DGCNN)* & \checkmark                   & -                    & 1.9    & 93.4 \textcolor{red}{(+0.5)} \\  \bottomrule
\end{tabular*}
\label{table:modelnet_cls}
\vspace{-2mm}
\end{table}

\textbf{3D object classification on ModelNet40.} We conducted classification experiments on the ModelNet40~\cite{ModelNet40} dataset, which comprises 9,843 training samples and 2,468 testing samples. Our approach utilizes 1,024 points as input and does not employ a voting strategy. The results are presented in Table~\ref{table:modelnet_cls}. Our proposed module enhances the performance of PointNet++ and DGCNN by 1.8\% and 1.1\% respectively. It achieves superior performance compared to existing Transformer-based architectures while maintaining a low parameter count. Besides, we replace the additional MLP overhead in the attention heads with the base network's feature extraction layer, achieving improvements of 1.4\% and 0.8\% in base networks. We also introduce a lightweight elite version, which replaces multi-head attention with single-head attention and incorporates a low-rank decomposition parameter set to $r$ = 16. This elite version maintains competitive performance with other architectures. We evaluated the performance under PointMLP, where it approaches 94.5\% in accuracy. This demonstrates that our method can still achieve effective improvements even on models with larger parameter sizes. Moreover, we observed that the lightweight version outperforms due to the superior feature extraction of PointMLP. Overall, our module offers flexible and principled trade-offs between computational efficiency and model performance.

We further explored our pre-train model, where the pre trained  network parameters are frozen, and only the linear classification layer is trained. We choose DGCNN as backbone, and the results are presented in Table~\ref{table:froze}. Compared to baseline pre-training methods such as OcCo~\cite{occo}, which relies solely on masking, and CrossPoint~\cite{crosspoint}, which leverages contrastive learning, our pre-training approach consistently enables the network to better capture and represent point cloud features.

\begin{table}[t]
    \centering
    \begin{minipage}{0.29\textwidth}
        \renewcommand{\arraystretch}{1.2}
        \scriptsize
        \caption{Classification in ModelNet40, while training the linear classifier only.}
        \label{table:froze}
        \resizebox{\linewidth}{!}{
            \begin{tabular}{l c}
                \toprule
                Method & Acc. \\ \hline
                Point-BERT~\cite{Point-BERT} & 87.4 \\ 
                Point-MAE~\cite{Point-MAE}  & 91.0 \\ 
                Point-M2AE~\cite{Point-M2AE}  & 92.9 \\ 
                I2P-MAE~\cite{Zhang2022Learning3R} & \textbf{93.4} \\ \hline
                DHGCN~\cite{DHGCN} & 93.2 \\ 
                DGCNN-OcCo~\cite{occo} & 90.7 \\ 
                DGCNN-CrossPoint~\cite{crosspoint} & 91.2 \\ \hline
                Ours \textbf{w/o} GF-Core & 92.1 \\ 
                Ours \textbf{w} GF-Core & 92.7 \\ 
                \bottomrule
            \end{tabular}
        }
    \end{minipage}
    \hfill
    \begin{minipage}{0.68\textwidth}
        \renewcommand{\arraystretch}{1.2}
        \scriptsize
        \caption{Classification results on three variants of the ScanObjectNN dataset with DGCNN as the backbone, focusing on performance improvements and comparisons with advanced models.}
        \label{table:scanobjectnn}
        \resizebox{\linewidth}{!}{ 
            \begin{tabular}{l c c c c}
                \toprule
                Methods & Pre-trained & OBJ\_ONLY & OBJ\_BG & PB\_T50\_RS \\ \hline
                PointNet~\cite{pointnet} & - & 79.2 & 73.3 & 68.2 \\ 
                PointNet++~\cite{PointNet++} & - & 84.3 & 82.3 & 77.9 \\ 
                PointCNN~\cite{PointCNN} & - & 85.5 & 86.1 & 78.5 \\ 
                DGCNN~\cite{DGCNN} & - & 86.2 & 82.6 & 78.1 \\ \hline
                Point-BERT~\cite{Point-BERT} & \checkmark & 88.12 & 87.43 & 83.07 \\ 
                Point-MAE~\cite{Point-MAE}  & \checkmark & 88.29 & 90.02 & 85.18 \\ 
                Mamba3D~\cite{mamba3D} & \checkmark & 92.08 & 93.12 & 92.05 \\ 
                PointMamba~\cite{Pointmamba}  & \checkmark & 88.47 & 90.71 & 84.87 \\ \hline
                Ours & \checkmark & 89.16 \textcolor{red}{(+2.96)} & 88.64 \textcolor{red}{(+6.34)} & 84.42 \textcolor{red}{(+6.32)} \\ 
                \bottomrule
            \end{tabular}
        }
    \end{minipage}
\end{table}
\begin{table}[t]
    \centering
    \begin{minipage}{0.47\textwidth}
        \renewcommand{\arraystretch}{1.2}
        \scriptsize
        \caption{Classification results of ModelNet40 with limited training data, using DGCNN as the backbone.}
        \label{table:limited}
        \resizebox{\linewidth}{!}{
            \begin{tabular}{lccccc}
                \toprule
                Methods & \multicolumn{5}{c}{Limited training data ratios} \\
                \cmidrule(lr){2-6}
                & 1\% & 2\% & 5\% & 10\% & 20\% \\
                \midrule
                FoldingNet~\cite{Yang2017FoldingNetIU} & 56.4 & 66.9 & 75.6 & 81.2 & 83.6 \\ 
                MAE3D~\cite{MAE3D} & 61.7 & 69.2 & 80.8 & 84.7 & 88.3 \\ 
                DHGCN~\cite{DHGCN} & 62.7 & 72.2 & 81.3 & 86.1 & 89.1 \\ \hline
                Ours & \textbf{66.1} & \textbf{74.9} & \textbf{83.1} & \textbf{86.7} & \textbf{90.4} \\
                \bottomrule
            \end{tabular}
        }
    \end{minipage}
    \hfill
    \begin{minipage}{0.50\textwidth}
        \renewcommand{\arraystretch}{1.2}
        \scriptsize
        \caption{Part segmentation results on the ShapeNetPart dataset, using DGCNN as the backbone.}
        \label{table:segment}
        \resizebox{\linewidth}{!}{
            \begin{tabular}{l c c c }
                \toprule
                Methods & Param.(M) & Class mIoU & Instance mIoU \\
                \hline
                PointNet~\cite{pointnet} & 3.6 & 80.4 & 83.7 \\ 
                PointNet++~\cite{PointNet++} & 1.0 & 81.9 & 85.1 \\ 
                DGCNN~\cite{DGCNN}  & 1.3 & 82.3 & 85.2 \\ 
                PAConv~\cite{PAConv} & - & 84.2 & 86.0 \\ 
                Point-BERT~\cite{Point-BERT} & 27.1 & 84.1 & 85.6 \\ 
                Point-MAE~\cite{Point-MAE}  & 27.1 & 84.2 & \textbf{86.1} \\ 
                PointMamba~\cite{Pointmamba}  & 17.4 & \textbf{84.4} & 86.0 \\ \hline
                Ours & 1.5 & 83.9 \textcolor{red}{(+1.6)} & 85.7 \textcolor{red}{(+0.5)} \\ 
                \bottomrule
            \end{tabular}
        }
    \end{minipage}
\vspace{-2mm}
\end{table}
\textbf{Real-world object classification on ScanObjectNN.} The ScanObjectNN~\cite{scanobjectnn} dataset is a real-world 3D point cloud dataset containing 2,902 object instances across 15 categories, with three variants: OBJ\_ONLY, OBJ\_BG and PB\_T50\_RS. It poses challenges due to noise, background clutter, and occlusions, making it ideal for evaluating model robustness. The results are shown in Table~\ref{table:scanobjectnn}. Our model, integrated with DGCNN, achieves accuracy improvements of 2.96\%, 6.34\%, and 6.32\% across the three variants respectively. On the OBJ\_ONLY variant, our model surpasses most complex architectures. In more challenging variants, the GF-Core and pre-training strategy significantly enhance the backbone's robustness against noisy points and background interference. However, the backbone network architecture warrants further in-depth investigation to optimize performance in complex scenarios.

\textbf{Classification with limited data.} Following the previous work~\cite{DHGCN}, our pre-trained model achieves superior generalization under the limited training data and linear classifier traing only (Table~\ref{table:limited}). Our model surpasses existing benchmarks across all metrics and attains 66.1\% accuracy under the extreme 1\% data regime, demonstrating robust feature transferability.

\textbf{Part segmentation on ShapeNet Part.} The ShapeNet Part~\cite{shapenetpart} dataset is a widely used benchmark for 3D point cloud part segmentation, containing 16,881 shapes from 16 categories, with 2048 points per shape as input. We adopt average instance IoU and average class IoU as evaluation metrics. The results are presented in Table~\ref{table:segment}. Using DGCNN as the backbone, our approach achieves improvements of 1.6\% in class mIoU and 0.5\% in instance mIoU. Although our performance still lags behind other architectures, we accomplish this with less than 10\% of their parameters. Overall, these results demonstrate that foundational architectures like DGCNN retain significant potential for improvement, which is effectively harnessed by GF-Core.

\subsection{Ablation Study}
\textbf{Component ablation study.} We conducted an ablation study on the ModelNet40 dataset, where our model achieved great performance. Here, we decompose GF-Core into its Grouping part and Feature extraction part to evaluate their individual contributions to the backbone network. The results are shown in Table~\ref{table:component}. Replacing the simple max pooling in the feature extraction layer with a self-attention module, which better mitigates aggregation loss, significantly enhances the backbone's performance. Compared to the commonly used KNN, our fusion selection method—based on weight matrices and feature spaces—yields higher-quality point cloud neighborhoods, further improving performance. Additionally, the Grouping Perturbation pre-training strategy proves effective, delivering even greater gains when combined with GF-Core, aligning with our initial design expectations.

\textbf{Ratio of mask and noise in pre-train.} We employ the complex OBJ-BG variant (more relevant to our pretraining methodology) as test samples. The results are shown in Table.~\ref{table:mask}. By adjusting the masking and noise ratio, we achieved a 2.75\% performance improvement. Under high perturbation conditions, the feature consistency learning task becomes more challenging, enabling our model to capture and aggregate more discriminative features without dataset modifications. 
\begin{wrapfigure}{r}{6.5cm}
    \centering
    \vspace{-4mm}
    \hspace{-5mm}
    \includegraphics[width=0.49\textwidth]{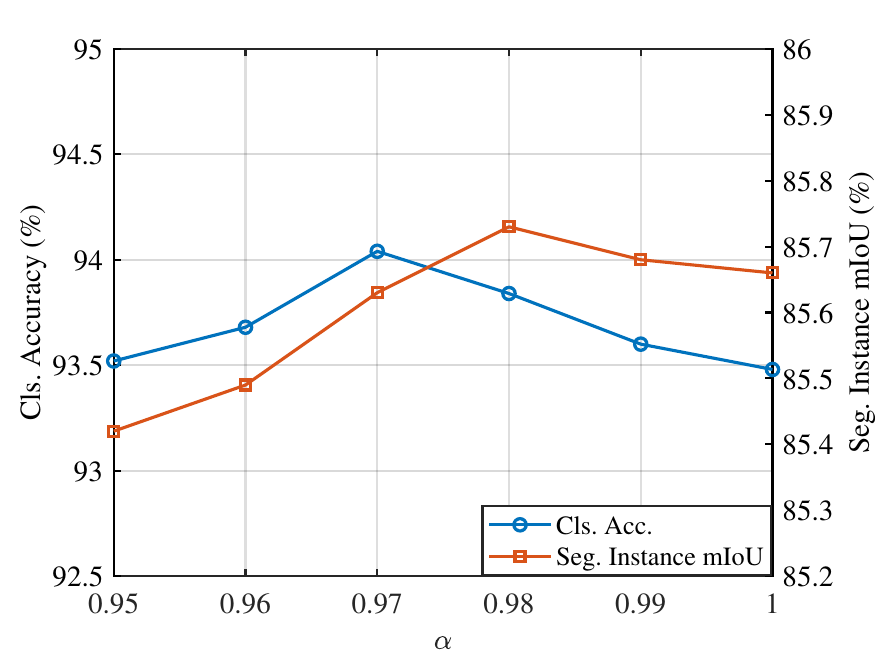}
    \vspace{-6mm}
    \caption{Ablation results on different $\alpha$ in pre-train.}
    \vspace{-10mm}
    \label{fig:alpha}
\end{wrapfigure}
It should be noted that excessive perturbation may compromise information extraction from point clouds, potentially degrading model performance.

\begin{table}[t]
    \centering
    \begin{minipage}{0.54\textwidth}
        \renewcommand{\arraystretch}{1.2}
        \scriptsize
        \caption{Component ablation study in ModelNet40 dataset, using GF-Core to replace max pooling and KNN in the extraction and grouping layer and simultaneously verifying the effectiveness of pre training.}
        \label{table:component}
        \resizebox{\linewidth}{!}{
            \begin{tabular}{cccc}
                \toprule
                Extraction & Grouping & Pre-train & Acc. \\
                \hline
                x & × & × & 92.9 (DGCNN) \\
                × & × & \checkmark & 93.1 \textcolor{red}{(+0.2)} \\
                \checkmark & × & × & 93.5 \textcolor{red}{(+0.6)} \\
                \checkmark & \checkmark & × & 93.7 \textcolor{red}{(+0.8)} \\
                \checkmark & \checkmark & \checkmark & 94.0 \textcolor{red}{(+1.1)} \\
                \bottomrule
            \end{tabular}
        }
    \end{minipage}
    \hfill
    \begin{minipage}{0.44\textwidth}
        \renewcommand{\arraystretch}{1.2}
        \scriptsize
        \caption{Classification results on OBJ-BG variants. Experiment on ablation by changing the number of mask and noise points in the four layer backbone network in pre-train.}
        \label{table:mask}
        \resizebox{\linewidth}{!}{
            \begin{tabular}{c c c}
                \toprule
                Mask Num. & Noise Num. & Acc. \\
                \hline
                (1,2,3,4) & (1,2,3,4) & 85.89 \\
                (1,2,3,4) & (2,4,6,8) & 86.57 \\
                (2,4,6,8) & (2,4,6,8) & 87.44 \\
                (6,10,12,16) & (2,4,6,8) & \textbf{88.64} \\
                (15,15,15,15) & (2,4,6,8) & 86.75 \\
                \bottomrule
            \end{tabular}
        }
    \end{minipage}
\vspace{-3mm}
\end{table}

\textbf{Global and local trade-off in pre-train.} We conducted classification and segmentation benchmarks on ModelNet40 and ShapeNet Part datasets respectively, using DGCNN as the backbone. The results are shown in Fig.~\ref{fig:alpha}. The task of classification achieves optimal performance with smaller $\alpha$ values compared to segmentation, as it inherently imposes stronger constraints on global feature aggregation. This empirical validation demonstrates the effectiveness of our feature trade-off mechanism between local and global representations.


\section{Conclusion}
We presented GF-Core, a lightweight module that revitalized classical point cloud networks by harmonizing neighborhood grouping and feature aggregation through attention-guided coordination. By preserving the simplicity of frameworks like DGCNN and PointNet, we addressed their underutilized potential. A self-supervised pretraining strategy further enhanced robustness to point-level feature learning ability. Our work demonstrated that strategic refinement of foundational architectures, rather than structural complexity, could also drive progress in 3D vision. In future work, we will explore adaptive grouping size strategies and broader use of GF-Core across diverse point cloud tasks.

{\small
\bibliographystyle{plain}
\bibliography{reference}
}

\newpage
\appendix

\section{Appendix}
\subsection{Self-Supervised Pretraining via Grouping
Perturbation details}
\begin{figure}[H]
    \centering
    \includegraphics[width=1\linewidth]{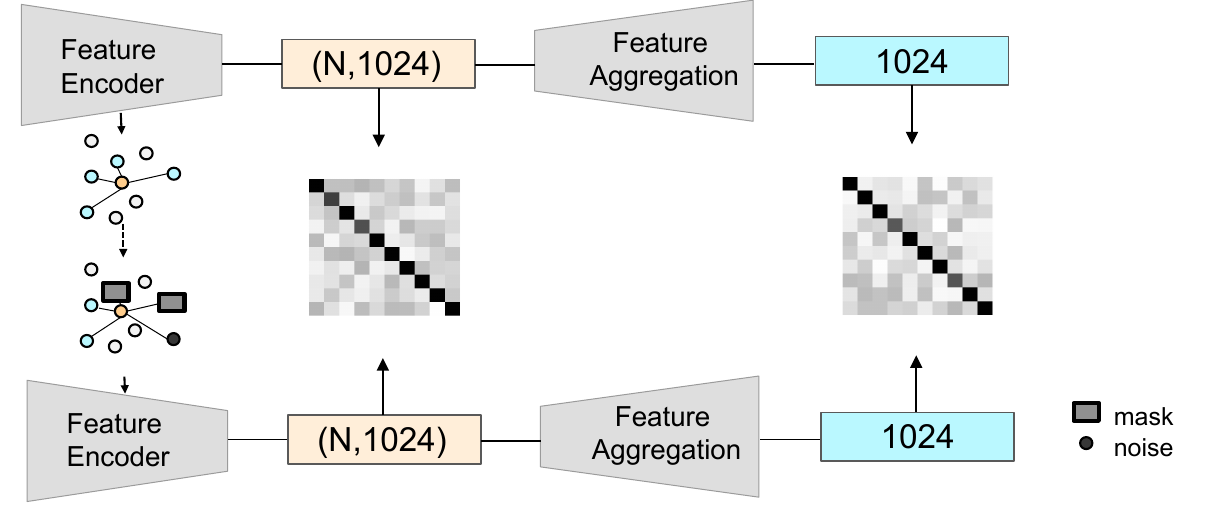}
	\caption{The details of our Self-Supervised Pretraining}
	\label{fig:pretrain}
\end{figure}
The overall pre-training pipeline is illustrated in Fig.~\ref{fig:pretrain}. At its core, our GF-Core module is designed to fully exploit the rich relational information among points in a point cloud, with a particular emphasis on neighborhood interactions. Additionally, we aim for our approach to unlock the full potential of the backbone network. Guided by these two principles, our self-supervised learning framework is built to extract hidden dependencies between points through neighborhood perturbation and contrastive learning, and is intended to be broadly applicable to point-based architectures.

Our method draws inspiration from existing point cloud pre-training approaches. On one hand, after grouping is performed, we deliberately mask a subset of neighboring points. Notably, in reference to methods such as Point-MAE~\cite{Point-MAE}, we mask only the feature information while retaining the geometric coordinates. This strategy promotes the learning of latent geometric and feature dependencies along with effective feature representations from the limited neighboring information and geometric context. On the other hand, we introduce several irrelevant distractor points into each group, enabling GF-Core to robustly learn point-to-point relationships even in the presence of grouping errors, thereby enhancing its generalization capability and ability to learn attention weights.

This design achieves three-way alignment: 1) Spatial robustness through controlled neighborhood perturbation, 2) Semantic stability via cross-view correlation learning in Barlow Twins loss~\cite{barlowtwins}, and 3) Structural awareness through global and local constraints.

\subsection{Weight matrix and Fusion module details }
\begin{wrapfigure}[12]{r}{6cm}
    \centering
    \vspace{-4mm}
\includegraphics[width=0.40\textwidth]{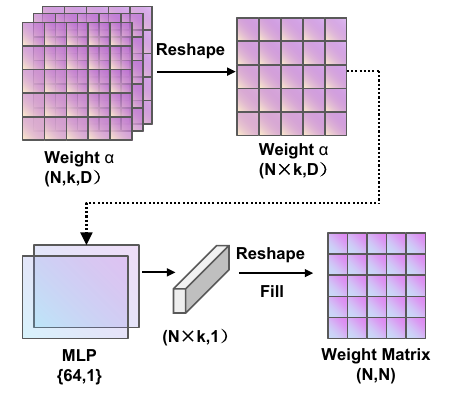}
    \caption{The details of Weight matrix}
    \label{fig:W}
\end{wrapfigure}

As shown in Fig.~\ref{fig:W}, we aggregate the D-dimensional attention weights using an MLP to obtain a scalar local weight for each neighbor with respect to the center point. Furthermore, since the selection of neighbors will change during subsequent grouping, we organize the attention weights of the N center points and their k neighbors into an $(N,N)$ matrix, where all entries corresponding to non-neighbor pairs are set to zero, indicating the absence of a neighbor relationship.

We further explore the impact of aligning the function in our Fusion module with that used in GATv2~\cite{GATv2}. We experiment with $\mathbf{a}^\top \mathrm{LeakyReLU}$ and $\mathrm{MLP}$ as the Fusion model respectively as follows:

\begin{equation}
\mathbf{D}_{combine1}^m = \mathbf{a}^\top \mathrm{LeakyReLU}\left([\mathbf{D}_{feature}^m||\mathbf{W}_{new} \right])
\label{eq:fusion2}
\end{equation}
\begin{equation}
\mathbf{D}_{combine2}^m = \mathrm{MLP}\left([\mathbf{D}_{feature}^m||\mathbf{W}_{new} \right])
\label{eq:fusion3}
\end{equation}
Intuitively, aligning the functions of the two modules could help reduce barriers to information flow. However, the core functions of the attention extraction and fusion modules are fundamentally different: while the attention module performs neighborhood-level attention analysis for each center point, the fusion module is designed to capture the global geometric and feature distributions among all points. Enforcing such alignment may therefore constrain the model’s expressive diversity. For this reason, we ultimately adopt an MLP as our fusion module.

\section{More experimental details}
\subsection{Details of different GF-Core versions}
GF-Core module is designed to enhance point-based architectures while maintaining architectural simplicity. To balance performance and lightweight, we tested different versions. 
\begin{equation}
\mathbf{G}_{independent} = \mathrm{MLP}(\mathbf{f}_{ij}) \in \mathbb{R}^{D} ,
\label{eq:parallel_process3}
\end{equation}
\begin{equation}
\mathbf{G}_{consistent} = \underbrace{\mathrm{Conv}(\mathbf{f}_{ij})}_{\mathclap{\mathrm{Base\ Network}}} \in \mathbb{R}^{D},
\label{eq:parallel_processcons}
\end{equation}
In our initial version, as shown in Eq.~\ref{eq:parallel_process3}, we employ a standalone MLP for feature preprocessing, which becomes increasingly costly as the number of points and feature dimensions grow. Fortunately, in conventional point-based architectures, both the network input and the input for our attention extraction module are based on relative features ($f_j - f_i$). This insight motivates our second version (Eq.~\ref{eq:parallel_processcons}), where we leverage the base network’s feature extraction layer for preprocessing, thereby reducing calculation costs while maintaining performance. Furthermore, we introduce an elite version by reducing both the number of attention heads and the parameterization of $a^\top$, achieving a more lightweight design. Empirically, all versions yield significant improvements over baseline architectures. We retain these variants to enable flexible trade-offs between computation costs and performance, thus facilitating effective deployment across a variety of architectures while ensuring robust generalization.

\subsection{Additional experiments}

\begin{table}[H]
    \centering
    \begin{minipage}{0.40\textwidth}
        \renewcommand{\arraystretch}{1.2}
        \scriptsize
        \caption{Classification results on ModelNet40 dataset across different values of k with DGCNN as the backbone.}
        \label{table:knn}
        \resizebox{\linewidth}{!}{ 
            \begin{tabular}{l c c c c}
                \toprule
                \multirow{2}{*}{Methods} & \multicolumn{4}{c}{k} \\
                \cmidrule(lr){2-5}
                & 5 & 10 & 20 & 40 \\
                \midrule
                DGCNN(backbone) & 90.5 & 91.4 & 92.9 & 92.4 \\ 
                Ours(w/o pretrain) & 91.3 & 92.0 & 93.7 & 93.3 \\ 
                Ours(w pretrain) & 91.9 & 92.4 & 94.0 & 93.9 \\ 
                \bottomrule
            \end{tabular}
        }
    \end{minipage}
    \hfill
    \begin{minipage}{0.58\textwidth}
        \renewcommand{\arraystretch}{1.2}
        \scriptsize
    \caption{Ablation study of inference speed on the ModelNet40 dataset,  with DGCNN as the backbone. We set the batch size to 16.}
        \label{table:inference}
        \resizebox{\linewidth}{!}{
            \begin{tabular}{ccccc}
    \toprule
    Extraction & Grouping &$\mathbf{G}_{use}$ & $Time_{batch}$(ms) & $Time_{sample}$(ms) \\
    \midrule
    × & × & - & 25.027 & 1.572 \\
    \checkmark & × & $\mathbf{G}_{ind.}$ & 46.418 & 2.922 \\
     \checkmark &\checkmark  & $\mathbf{G}_{ind.}$ & 53.952 & 3.380 \\
     \checkmark &\checkmark  & $\mathbf{G}_{con.}$ & 38.443 & 2.462 \\
    \bottomrule
            \end{tabular}
        }
    \end{minipage}
\end{table}
\ptitle{Ablation Study on Different Values of k.}
Both our GF-Core module and pretraining strategy are closely tied to the neighborhood structure of each center point, making the number of neighbors (k) a critical factor for feature extraction. Following the parameter choices in DGCNN, we evaluate classification performance on ModelNet40~\cite{ModelNet40} across different values of k, as shown in Table~\ref{table:knn}. We observe that increasing k within a reasonable range allows for richer feature aggregation, whereas excessively large k introduces irrelevant neighbors and ultimately degrades performance. Besides, by virtue of our grouping perturbation pretraining, the model maintains robust performance even with substantially increased k-values.
This analysis also points to an interesting direction: adaptively selecting the optimal k for each layer as well as each point cloud could further boost performance. We leave this promising avenue for future exploration.

\ptitle{Inference Speed Evaluation.} We evaluated the inference speed on the ModelNet40 test set, as reported in Table~\ref{table:inference}. By leveraging the computational efficiency of backbone networks like DGCNN, our method achieves practically viable inference speeds despite introducing latency overhead. Furthermore, configuring different variants of $\mathbf{G}$ enables flexible accuracy-latency trade-offs, adapting our framework to accommodate diverse application requirements. 
\begin{wraptable}{r}{0.25\columnwidth}
  \centering
  \vspace{-3mm}
  \small
  \setlength{\tabcolsep}{8pt}
  \renewcommand{\arraystretch}{0.8}
  \begin{tabular}{cc}
    \toprule
    Head Num. & Acc. \\
    \midrule
    1 & 93.6 \\
    2 & 94.0 \\
    3 & 94.2 \\
    4 & 94.2 \\
    \bottomrule
  \end{tabular}
  \caption{Effect of the number of attention heads.}
  \label{table:headnum}
  \vspace{-3mm}
\end{wraptable}

\ptitle{Ablation Study on the Number of Attention Heads.} We conducted experiments on ModelNet40 to characterize the impact of attention head count on model performance. As shown in Table~\ref{table:headnum}, increasing the number of heads from one to multiple leads to clear improvements in network performance. However, marginal returns diminish significantly despite escalating computational costs. Considering the additional caculation costs and latency introduced by more heads, we selected two attention heads as the default setting for our other experiments.


\end{document}